\documentclass{tlp}

\usepackage[utf8]{inputenc}
\usepackage{amsmath}
\usepackage{amssymb}
\usepackage{microtype}
\usepackage{url}
\usepackage{graphicx}
\usepackage{xcolor}

\usepackage{listings}
\providecommand{\Underscore}{\textunderscore}

\lstdefinelanguage{clingo}{basicstyle=\ttfamily,keywordstyle=[1]\bfseries,keywordstyle=[2]\bfseries,keywordstyle=[3]\bfseries,showstringspaces=false,literate={_}{\Underscore}1 {\%\%}{}0,escapeinside={\#(}{\#)},alsoletter={\#,\&},keywords=[1]{not,from,import,def,if,else,elif,return,while,break,and,or,for,del,and,class,with,as,is,yield,async},keywords=[2]{\#const,\#show,\#minimize,\#base,\#theory,\#count,\#external,\#program,\#script,\#end,\#heuristic,\#edge,\#project,\#show,\#sum},keywords=[3]{&,&dom,&sum,&diff,&show},morecomment=[l]{\#\ },morecomment=[l]{\%\ },morestring=[b]",stringstyle={\itshape},commentstyle={\color{darkgray}}}

\lstdefinelanguage{python}{basicstyle=\ttfamily,keywordstyle=[1]\bfseries,showstringspaces=false,literate={_}{\Underscore}{1},escapeinside={\#(}{\#)},alsoletter={\#,\&},keywords=[1]{not,from,import,def,if,else,elif,return,while,break,and,or,for,in,del,and,class,with,as,is,yield,async},morecomment=[l]{\#\ },morestring=[b]",stringstyle={\itshape},commentstyle={\color{darkgray}}}
 \sbox0{\ttfamily A}
\edef\mybasewidth{\the\wd0 }
\sbox0{\scriptsize\ttfamily A}
\edef\mybasewidths{\the\wd0 }
\lstdefinelanguage{clingoht}{language=clingo,
  columns=flexible,
  basewidth=\mybasewidth,
  escapeinside=||,
  mathescape,
  float=ht
}
\lstdefinelanguage{clingos}{language=clingo,basicstyle=\small\ttfamily }

\providecommand{\sysfont}{\textit}

\newcommand{\clinguin}{\sysfont{clinguin}}

\newcommand{\clingo}{\sysfont{clingo}}

\newcommand{\powerset}[1]{\ensuremath{2^{#1}}}
\newtheorem{example}{Example}

\newcommand{\sebfunc}{\ensuremath{a}}

\lefttitle{Susana Hahn et al.}

\jnlPage{1}{8}
\jnlDoiYr{2021}
\doival{10.1017/xxxxx}

\begin{document}

\title[Reasoning about Study Regulations in ASP]{Reasoning about Study Regulations\\ in Answer Set Programming\thanks{A preliminary version of this article appeared in~\citep{hamaneotroscsc23a}.}}

\begin{authgrp}
  \author{\sn{Susana} \gn{Hahn}}
  \affiliation{University of Potsdam, Germany}\affiliation{Potassco Solutions, Germany}
  \author{\sn{Cedric} \gn{Martens}}
  \affiliation{University of Potsdam, Germany}
  \author{\sn{Amade} \gn{Nemes}}
  \affiliation{University of Potsdam, Germany}
  \author{\sn{Henry} \gn{Otunuya}}
  \affiliation{University of Potsdam, Germany}
  \author{\sn{Javier} \gn{Romero}}
  \affiliation{University of Potsdam, Germany}
  \author{\sn{Torsten} \gn{Schaub}}
  \affiliation{University of Potsdam, Germany}\affiliation{Potassco Solutions, Germany}
  \author{\sn{Sebastian}~\gn{Schellhorn}}
  \affiliation{University of Potsdam, Germany}
\end{authgrp}

\maketitle

\begin{abstract}
  We are interested in automating reasoning with and about study regulations,
  catering to various stakeholders, ranging from administrators, over faculty, to students at different stages.
  Our work builds on an extensive analysis of various study programs at the University of Potsdam.
  The conceptualization of the underlying principles provides us with a formal account of study regulations.
  In particular, the formalization reveals the properties of admissible study plans.
  With these at end,
  we propose an encoding of study regulations in Answer Set Programming that produces corresponding study plans.
  Finally, we show how this approach can be extended to a generic user interface for exploring study plans.
\end{abstract}
 \begin{keywords}
   Answer Set Programming, Study regulations and plans
\end{keywords}
\section{Introduction}\label{sec:introduction}

Study regulations govern our teaching at universities
by specifying requirements to be met by students to earn a degree.
Creating a study program involves different stakeholders:
faculty members designing study programs,
administrative and legal staff warranting criteria, like studyability,
faculty members teaching the corresponding programs as well as supervising their execution on examination boards,
study advisors consulting students,
and of course, students studying accordingly.

Given this impressive spectrum of use-cases,
it is quite remarkable that study regulations are usually rather sparse and
leave many aspects to the commonsense of the respective users.
This is needed to cope with their inherent incomplete, inconsistent, and evolving nature.
For instance,
often study regulations leave open minor dependencies among modules.
Sometimes associated courses overlap and certain modules cannot be taken in the same semester.
And finally, studying happens over time, students' perspectives may change and faculty may rotate.
Often these phenomena are compensated by changes, preferences, recommendations, defaults, etc.
In fact,
this richness in issues and notions from Knowledge Representation and Reasoning (KRR)
makes study regulations a prime candidate for a comprehensive benchmark for KRR formalisms.

Our approach is adaptable, though not universally applicable.
Its foundations lie in the principles of Europe's Bologna Process,
designed to harmonize higher education across the continent.
European curricula, for example, are structured around credit points, often allocated to individual modules.
The European Credit Transfer and Accumulation System (ECTS) provides guidance on program design and credit allocation.
Similarly, our approach utilizes modules and credit points as fundamental building blocks.
Our goal is to encompass all study regulations at the University of Potsdam, which share additional key principles.
However, adaptation or even redesign may be necessary for other institutions with differing structures or requirements.

In fact,
this work is part of a project conducted at the University of Potsdam
to assist different users by automatizing study regulations.
These users range from study administrators, over faculty in different functions,
to prospective and advanced students.
We started by analyzing more than a dozen different study regulations in order to identify their underlying principles.
The conceptualization of the basic principles led us to a formal account of basic study regulations,
presented in Section~\ref{sec:approach}.
For illustration, we provide the formalization of the master program \emph{Cognitive Systems}.
We refine this in Section~\ref{sec:examination} by
showing how modules are passed by accomplishing their associated examination tasks.
The formalization of study regulations reveals the properties of admissible study plans.
To automatize reasoning about study regulations and their study plans,
we capture their properties in Answer Set Programming (ASP;~\cite{lifschitz02a}),
a declarative problem solving paradigm,
tailored for knowledge representation and reasoning.
The ASP-based encoding of basic study regulations is discussed in Section~\ref{sec:encodings}.
Moreover, we show in Section~\ref{sec:ui} how this encoding can be used together with an ASP-driven user interface
to browse through study plans of given study regulations.
We conclude in Section~\ref{sec:discussion}.
 \section{Conceptualizing study regulations}\label{sec:approach}

The basic concept of our study regulations are modules.
Accordingly,
a semester is composed of a set of modules and
a study plan is a finite sequence of semesters.
More formally,
given a set $M$ of modules,
a \emph{study plan} of $n$ semesters is a sequence
\(
(S_i)_{i=1}^n
\)
where
\(
S_i\subseteq M
\)
for $1\leq i\leq n$.
Study regulations specify legal study plans.
To capture this,
we propose an abstract characterization of study regulations and
show how they induce legal study plans.

A basic \emph{study regulation} is a tuple
\(
(M, G, c, s, l, u, R_g, R_t),
\)
where
\begin{enumerate}
\item $M$ is a set of modules,
\item $G\subseteq\powerset{M}$ distinguishes certain groups of modules,
\item $c: M \to \mathbb{N}$ gives the credits of each module,
\item $s: M \to \{w,s,e\}$ assigns a regular semester to a module,
\item $l: G \to \mathbb{N}$ returns the lower-bound of the credits of a module group,
\item $u: G \to \mathbb{N}$ returns the upper-bound of the credits of a module group,
\item $R_g \subseteq \powerset{(\powerset{M})^n}$ is a set of global   constraints expressing study regulations, and
\item $R_t \subseteq \powerset{(\powerset{M})^n}$ is a set of temporal constraints expressing study regulations.
\end{enumerate}
The module groups in $G$ allow us to structure the modules and to express group-wise regulations.
Functions $c$ and $s$ give the credit points of a module and its turnus,\footnote{Winter and summer semesters are associated with odd and even positions in a sequence, respectively (see below).}
viz.\ in winter, summer, or each semester (indicated by  \textit{w}, \textit{s}, \textit{e}), respectively.
The elements $l$ and $u$ are partial functions delineating the number of credits obtained per module group;
a specific number of credits is captured by an equal lower and upper bound.
The regulation constraints in $R_g$ and $R_t$ are represented extensionally:
each constraint $r$ in $R_g \cup R_t$ is represented by the sequences of sets of modules $r\subseteq(\powerset{M})^n$ satisfying it.
While the sets of constraints $R_g$ and $R_t$ share the same mathematical structure, they differ in purpose and are therefore separated for clarity's sake.
$R_t$ expresses temporal constraints over the sequence of semesters,
while $R_g$ does not make use of this sequential structure, and rather expresses global constraints over the entire set of modules.

A study plan for a basic study regulation
is a finite sequence of sets of modules satisfying all regulation constraints.
More precisely,
a sequence
\(
(S_i)_{i=1}^n
\)
of modules of length $n$
such that $S_i\subseteq M$ for $1\leq i\leq n$
is a \emph{study plan} for $(M, G, c, s, l, u, R_g, R_t)$ if
\(
(S_i)_{i=1}^n\in\bigcap_{r\in R_g \cup R_t} r
\).

Finally,
we call modules exogenous if they are imposed by external means, e.g.\ by an examination board.
The specific choice of these modules is determined case by case.

 \begin{example}[Cognitive systems]\label{sec:example:cogsys}

As an example,
consider the study regulations of the international master program \emph{Cognitive Systems}
offered at the University of Potsdam.\footnote{Available at \url{https://www.uni-potsdam.de/fileadmin/projects/studium/docs/03_studium_konkret/07_rechtsgrundlagen/studienordnungen/StO_CogSys_EN.pdf}}
This program offers a combination of modules in
Natural language processing,
Machine learning, and
Knowledge representation and reasoning.

These subjects are reflected by the three mandatory base modules, joined in $B$ in~\eqref{cogsys:base:modules}.
Since each module yields 9 credits,
their obligation is achieved by requiring that the modules stemming from $B$ must account for 27 credits.\footnote{This is not our way of modeling mandatory courses but rather reflects the actual regulation.}
While the amount is specified in~\eqref{cogsys:group:modules:lower} and~\eqref{cogsys:group:modules:upper},
the actual constraint is imposed in~\eqref{cogsys:group:modules:cardinality}.
The same constraint is used for choosing two among three possible project modules from~$P$.
The optional modules in group $O$ are handled similarly,
just that only 24 credits from 36 possible credits are admissible.
That is, four out of nine modules must be taken.
The freedom of which four the student may choose is restricted by the examination board by
imposing the study of up to two foundational modules $E\subseteq F$, which must be the only modules taken from module group $F$, as formalized in constraint~\eqref{cogsys:exogenous}.
The total number of credits over all modules must equal 120.
Finally, an internship, \textit{im}, and a thesis, \textit{msc}, are imposed in~\eqref{cogsys:msc}.
This brings us to the temporal regulation in~\eqref{cogsys:msc:pre} requiring that
at least 90 credits are accumulated before conducting a thesis.
The temporal regulations in~\eqref{cogsys:winter} and~\eqref{cogsys:summer} ensure that modules are taken in the right season.
And finally~\eqref{cogsys:disjoint} makes sure that modules are chosen at most once.

Let $E\subseteq F$ be some exogenous set of modules in the following example;
and let
$\overline{S}$ stand for $\bigcup_{i=1}^n S_i$, and
$M_w=\{m\in M\mid s(m)=w\}$ and $M_s=\{m\in M\mid s(m)=s\}$.
Then, the study regulations of the master program \emph{Cognitive Systems} with respect to $E$ can be formalized as follows.
\begin{align}
  M&=B \cup F \cup A \cup P \cup \{\mathit{im},\mathit{msc}\}
  \\
  G&={\{B,F,A,O,P,M\}}
  \\\label{cogsys:base:modules}
  &\qquad{B}=\{\mathit{bm}_i\mid i=1..3\}
  \\
  &\qquad{F}=\{\mathit{fm}_i \mid i=1..3\}
  \\
  &\qquad{A}=\{\mathit{am}_{i,j}\mid i=1..3,j=1,2\}
  \\
  &\qquad{O}=F \cup A
  \\
  &\qquad{P}=\{\mathit{pm}_i\mid i=1..3\}
  \\
  c&=\{\mathit{bm}_i\mapsto 9\mid i=1..3\}
     \cup{}
     \{\mathit{am}_{i,j}\mapsto 6\mid i=1..3,j=1,2\}
     \,\cup
     \\&\quad\
     \{\mathit{fm}_i \mapsto 6 \mid i=1..3 \}
     \cup{}
     \{\mathit{pm}_i\mapsto 12\mid i=1..3\}
     \cup{}
     \{\mathit{im}\mapsto 15,\mathit{msc}\mapsto 30\}
  \\
  s&=\{\mathit{bm}_1\mapsto w, \mathit{bm}_2\mapsto s, \mathit{bm}_3\mapsto w\}
     \cup{}
     \{\mathit{am}_{i,j}\mapsto e\mid i=1..3,j=1,2\}
     \,\cup
     \\&\quad\
     \{\mathit{fm}_i\mapsto w\mid i=1..3 \}
     \cup{}
     \{\mathit{pm}_i\mapsto e\mid i=1..3\}
     \cup{}
     \{\mathit{im}\mapsto e,\mathit{msc}\mapsto e\}
  \\\label{cogsys:group:modules:lower}
  l&=\{{B}\mapsto 27, {O}\mapsto 24, {P}\mapsto 24, {M}\mapsto 120 \}
  \\\label{cogsys:group:modules:upper}
  u&=\{{B}\mapsto 27, {O}\mapsto 24, {P}\mapsto 24, {M}\mapsto 120 \}
  \\
  R_g&=\{\,
     \label{cogsys:exogenous}
     \{(S_i)_{i=1}^n\subseteq M^n\mid |E| \leq 2, \overline{S} \cap F = E\},
     \\\label{cogsys:group:modules:cardinality}&\qquad
     \{(S_i)_{i=1}^n\subseteq M^n\mid l(H) \leq \textstyle\sum_{m\in H\cap \overline{S}}c(m) \leq u(H) \}
     \textrm{ for all } H \in \{B,O,P,M\},
     \\\label{cogsys:msc}&\qquad
     \{(S_i)_{i=1}^n\subseteq M^n\mid \{\mathit{im},\mathit{msc}\}\subseteq \overline{S}\}
     \,\}
  \\
  R_t&=\{\,
       \label{cogsys:disjoint}
       \{(S_i)_{i=1}^n\subseteq M^n\mid S_i\cap S_j =\emptyset,1\leq i<j \leq n\}
       \\\label{cogsys:winter}&\qquad
       \{(S_i)_{i=1}^n\subseteq M^n\mid M_w\cap S_{2k} =\emptyset,1\leq 2k  \leq n,k\in\mathbb{N}\}
       \\\label{cogsys:summer}&\qquad
       \{(S_i)_{i=1}^n\subseteq M^n\mid M_s\cap S_{2k-1}=\emptyset,1\leq 2k-1\leq n,k\in\mathbb{N}\}
       \\\label{cogsys:msc:pre}&\qquad
       \{(S_i)_{i=1}^n\subseteq M^n\mid \mathit{msc}\in S_k, \textstyle\sum_{1\leq i<k}\sum_{m\in S_i}c(m)\geq 90,k\in\mathbb{N}\}
     \,\}
\end{align}
If the set of exogenous modules given by the examination board is, for example, $E=\{\mathit{fm}_{1}\}$, one admissible study plan spanning four semesters is $S = (S_i)_{i=1}^4$, where
\begin{align}
  \label{ex:study:plan:one}
  S_1&=\{\mathit{bm}_1,\mathit{bm}_3,\mathit{fm}_{1},\mathit{am}_{1,2}\}
  \\\label{ex:study:plan:two}
  S_2&=\{\mathit{bm}_2,\mathit{am}_{2,1},\mathit{pm}_1\}
  \\\label{ex:study:plan:tri}
  S_3&=\{\mathit{im},\mathit{pm}_3,\mathit{am}_{3,1}\}
  \\\label{ex:study:plan:for}
  S_4&=\{\mathit{msc}\}
\end{align}
This plan comprises 120 credits, although the load per semester varies.

For illustration,
let us verify that our study plan belongs to the ones in~\eqref{cogsys:exogenous} and \eqref{cogsys:group:modules:cardinality} for $H = O$.
Indeed constraint~\eqref{cogsys:exogenous} is satisfied as we have $F \cap \overline{S} = \{\mathit{fm}_{1}\} = E$, and
thus $S$ is an element of constraint~\eqref{cogsys:exogenous}.
With regards to \eqref{cogsys:group:modules:cardinality}, we have
\begin{align}
  \label{ex:cogsys:A:cardinality:one}
  O \cap\overline{S}&=\{\mathit{fm}_{1},\mathit{am}_{1,2},\mathit{am}_{2,1},\mathit{am}_{3,1}\}
\end{align}
which makes us check whether our study plan satisfies
\begin{align}
  \label{ex:cogsys:A:cardinality:two}
  l(O)=24\leq \textstyle\sum_{m\in\{\mathit{fm}_{1},\mathit{am}_{1,2},\mathit{am}_{2,1},\mathit{am}_{3,1}\} }c(m) \leq 24= u(O)
\end{align}
This is indeed the case since
\(
c(\mathit{fm}_{1})+c(\mathit{am}_{1,2})+c(\mathit{am}_{2,1})+c(\mathit{am}_{3,1})=24
\).
Hence, our study plan is an element of constraint ~\eqref{cogsys:group:modules:cardinality}.
\end{example}

Although the above specification reflects the legal study regulation,
it leaves many ambiguities behind.
For instance, the number of credits per semester is left open,
as is the order of the modules.
The guideline is usually to take around 30 credits per semesters but this is not enforced.
Similarly, basic modules in $B$ should be taken before advanced ones in $A$,
again this is neither enforced nor always possible.
Since these constraints are usually soft, they are left to the students and/or their study advisors.
 \par
Our previous preliminary work~\citep{hamaneotroscsc23a} outlined further extensions to basic study regulations
(subarea specializations, module dependencies, blocking modules).
The following section delves into a more novel aspect:
how modules are passed by accomplishing their associated examination tasks.

\section{Examination Tasks}
\label{sec:examination}

Modules have specific examination tasks that define the criteria for successful completion.
Students typically complete these tasks within courses,
where they are carefully designed to align with the requirements of specific modules.

Study regulations focus on modules and their associated examination tasks.
It is the responsibility of each department to provide a course selection that enables students to complete the modules
required by their study program.
Therefore, we concentrate in what follows on modules and their associated examinations, rather than courses.

Each module has at least one examination task\footnote{All this follows the general regulations for study examinations for BSc and MSc degrees at
  \url{http://www.uni-potsdam.de/am-up/2013/ambek-2013-03-035-055.pdf} (last accessed on 30th of April 2024).}
and these tasks are unique to each module.
We distinguish between
\emph{primary examinations}   (e.g.\ written or oral exams) and
\emph{secondary examinations} (e.g.\ weekly exercises),
and denote them by $E_{p}$ and $E_{s}$, respectively.
In analogy to study plans,
\emph{examination plans} are sequences of form
\(
(E_i)_{i=1}^n
\subseteq
(E_p \cup E_s)^n
\).

Interestingly,
a single module can offer flexibility through different combinations of primary and secondary examination tasks.
Given a set of modules $M$ and sets $E_{p}$ and $E_{s}$ of primary and secondary examinations, we define the functions
$e_p: M \rightarrow \powerset{\powerset{E_p}}$
and
$e_s: M \rightarrow \powerset{\powerset{E_s}}$
to associate a module with different combinations of primary and secondary examination tasks, respectively.

For illustration,
suppose module $\mathit{bm}_1$ requires as primary examination task either a written exam or a final project report,
identified by $\mathit{ep}_{\mathit{bm}_1,1}$ and $\mathit{ep}_{\mathit{bm}_1,2}$.
Also,
two secondary examination tasks, a lecture attendance of 50\% and the successful completion of all weekly exercises,
viz.\ $\mathit{es}_{\mathit{bm}_1,1}$ and $\mathit{es}_{\mathit{bm}_1,2}$,
are required by the module.
This is captured as
$e_p(\mathit{bm}_1) = \{\{\mathit{ep}_{\mathit{bm}_1,1}\}, \{\mathit{ep}_{\mathit{bm}_1,2}\}\}$ and
$e_s(\mathit{bm}_1) = \{\{\mathit{es}_{\mathit{bm}_1,1},     \mathit{es}_{\mathit{bm}_1,2}\}\}$.

Next, we define the completion of modules in terms of examinations.

To this end,
we define a function \sebfunc\ associating examination plans with sequences of modules as
$\sebfunc: (E_i)_{i=1}^{n} \mapsto (S_i)_{i=1}^n$
where for $1\leq i\leq n$ we have
\[\textstyle
  S_i = \{ m\in M \mid
  V\cup W \subseteq \bigcup_{j=1}^{i}E_j, V\cup W \not\subseteq \bigcup_{j=1}^{i-1}E_j
  \text{ for some }V \in e_s(m), W \in e_p(m)
  \}
\]
The idea is to complete modules as early as possible.
Once a module is completed for a specific $V$ and $W$ combination,
it cannot be repeated for credit in later semesters.
A module may appear multiple times in a module sequence if it offers several valid combinations of primary and secondary examination tasks, and these combinations are spread across different semesters within the examination plan.
However,
study plans become invalid if they include the same module in multiple semesters, as this violates a core regulation
(see also~\eqref{cogsys:disjoint}).
Also,
different examination plans may result in the same module sequence,
especially if some examination tasks do not immediately contribute to completing a module.

For illustration,
let us continue our example with $e_p(\mathit{bm}_1)$ and $e_s(\mathit{bm}_1)$ as above.
Furthermore, suppose we have the primary examination tasks
\[
  e_p(\mathit{m})=\{\{ep_{\mathit{m},1}\}\}
  \text{ for }
  m\in \{\mathit{bm}_3, \mathit{fm}_1, \mathit{am}_{1,2}, \mathit{bm}_2, \mathit{am}_{2,1}, \mathit{pm}_1, \mathit{im}, \mathit{pm}_3, \mathit{am}_{3,1}, \mathit{msc}\}
\]
along with the following secondary examination tasks
\(
e_s(\mathit{bm}_2)=\{\{es_{\mathit{bm}_2,1}\}\}
\),
\(
e_s(\mathit{bm}_3)=\{\{es_{\mathit{bm}_3,1}, es_{\mathit{bm}_3,2}\}\}
\),
\(
e_s(\mathit{fm}_1)=\{\{es_{\mathit{fm}_1,1}\}\}
\),
and
\[
  e_s(\mathit{m})=\{\emptyset\}
  \text{ for }
  m\in \{\mathit{am}_{1,2}, \mathit{am}_{2,1}, \mathit{pm}_1, \mathit{im}, \mathit{pm}_3, \mathit{am}_{3,1},\mathit{msc}\}.
\]

Now, consider the examination plan $(E_i)_{i=1}^{4}$ where
\begin{align*}
  E_1&=\{
       ep_{\mathit{bm}_1,1}, es_{\mathit{bm}_1,1}, es_{\mathit{bm}_1,2}, ep_{\mathit{bm}_3,1}, es_{\mathit{bm}_3,1}, es_{\mathit{bm}_3,2}, ep_{\mathit{fm}_1,1}, es_{\mathit{fm}_1,1}, ep_{\mathit{am}_{1,2},1} \}
  \\
  E_2&=\{
       ep_{\mathit{bm}_2,1}, es_{\mathit{bm}_2,1}, ep_{\mathit{am}_{2,1},1}, ep_{\mathit{pm}_1,1} \}
  \\
  E_3&=\{
       ep_{\mathit{im},1}, ep_{\mathit{pm}_3,1}, ep_{\mathit{am}_{3,1},1} \}
  \\
  E_4&=\{ep_{\mathit{msc}}\}.
\end{align*}
In fact,
this examination plan induces the study plan given in Section~\ref{sec:approach},
that is,
\(
\sebfunc((E_i)_{i=1}^4)=(S_i)_{i=1}^4
\)
as given in~\eqref{ex:study:plan:one} to~\eqref{ex:study:plan:for}.

Similarly, the examination plan $E'=(E_1, E_2\cup\{ep_{\mathit{bm}_1,2}\},E_3,E_4)$
induces $\sebfunc(E')=(S_1,S_2\cup\{\mathit{bm}_1\},S_3,S_4)$.
Since $ep_{\mathit{bm}_1,1}\in E_1$ and
$e_p(\mathit{bm}_1) = \{\{\mathit{ep}_{\mathit{bm}_1,1}\}, \{\mathit{ep}_{\mathit{bm}_1,2}\}\}$
indicates that only one primary examination tasks is needed to accomplish module $\mathit{bm}_1$,
the occurrence of $ep_{\mathit{bm}_1,2}$ in the second semester of $E'$ is redundant and
reflected by the second occurrence of module $\mathit{bm}_1$ in $\sebfunc(E')$.

The relation between primary and secondary examinations tasks is even more intricate since
secondary tasks may need to be accomplished before primary ones.
We represent such dependencies by a relation
\(
D \subseteq \powerset{\powerset{\mathit{E_s}}} \times \powerset{\mathit{E_p}}
\)
between
alternative sets of secondary examination tasks and sets of primary ones.\footnote{For simplicity, we refrain here from defining these relations in a module dependent way.
  Without loss of generality,
  we can thus assume that only examination tasks associated with the same module are put in correspondence.}
More precisely,
any such dependency in $D$ expresses a temporal constraint requiring that
one of the sets of secondary examination tasks must be accomplished
before or in the same semester as the set of primary constraints.

For example,
the dependencies
\(
(\{\{es_{\mathit{bm}_1,2}\}\}, \{ep_{\mathit{bm}_1,1}\} )
\)
and
\(
(\{\{es_{\mathit{bm}_1,2}\}\}, \{ ep_{\mathit{bm}_1,2}\})
\)
express that
weekly exercises of module $\mathit{bm}_1$ (viz.\ $es_{\mathit{bm}_1,2}$) must be successfully completed
before a student can take a written exam ($ep_{\mathit{bm}_1,1}$) or hand in the final project report ($ep_{\mathit{bm}_1,2}$).
None of the primary examinations depend on secondary examination task $es_{\mathit{bm}_1,1}$,
which is only needed to accomplish the module itself.

We are now ready to formalize the concept of an admissible examination plan.

A \emph{basic study regulation problem with examination tasks} is a pair $(\mathcal{B},\mathcal{E})$,
where
$\mathcal{B}$ is a basic study regulation problem
and
\(
\mathcal{E} = (E_p, E_s, e_p, e_s, D, R_{eg}, R_{et})
\)
where $E_p, E_s, e_p, e_s, D$ are as defined above, and
$R_{eg} \subseteq \powerset{(\powerset{E_p \cup E_s})^n}$ and
$R_{et} \subseteq \powerset{(\powerset{E_p \cup E_s})^n}$
are sets of global and temporal constraints on examination plans, respectively,
as given next.

In doing so, we extend the above abbreviation of $\overline{(E_i)_{i=1}^n}$ by $\overline{E}$
to sets of sets, that is,
\(
\overline{X}=\bigcup_{x\in X} x
\)
for any set $X$ of sets.
Now, given a set of modules $M$ along with primary and secondary examination tasks $E_p\cup E_s$,
we define the following constraints:
\begin{align}
  R_{eg} &=\{\label{module:completion:constraint:e1}
           \{
           (E_i)_{i=1}^n\subseteq (E_p \cup E_s)^n \mid
           \overline{E}\cap\overline{\mathit{e_p}(m)\cup\mathit{e_s}(m)}\neq\emptyset
           \text{ implies }\\
         &\label{module:completion:constraint:e2}\hspace{160pt}
           \overline{E}\cap\overline{\mathit{e_p}(m)}\in\mathit{e_p}(m)
           \text{ for all } m \in M
           \},\\
         &\label{module:completion:constraint:e3}
           \phantom{\ =\{}
           \{
           (E_i)_{i=1}^n\subseteq (E_p \cup E_s)^n \mid
          \overline{E}\cap\overline{\mathit{e_p}(m)\cup\mathit{e_s}(m)}\neq\emptyset
          \text{ implies }\\
         &\label{module:completion:constraint:e4}\hspace{160pt}
           \overline{E}\cap\overline{\mathit{e_s}(m)}\in\mathit{e_s}(m)
           \text{ for all } m \in M
           \}
           \}\\
  R_{et} &=\{\label{examination:uniqueness}
           \{
           (E_i)_{i=1}^n\subseteq (E_p \cup E_s)^n \mid E_i \cap E_j = \emptyset, 1 \leq i < j \leq n
           \},\\
         &\label{dependency:constraint:e1}
           \phantom{\ =\{}
           \{(E_i)_{i=1}^n\subseteq (E_p \cup E_s)^n \mid\text{for all }(X,W) \in D,\\
         &\label{dependency:constraint:e2}\hspace{50pt}
           W \subseteq \overline{E} \text{ implies there is some }V \in X \text{ such that }\\
         &\label{dependency:constraint:e3}\hspace{50pt}
           V \subseteq \overline{E} \text{ and }
          \max(\{i\mid V \cap E_i \neq \emptyset \}) \leq \min(\{i \mid W \cap E_i \neq \emptyset \})
           \}
           \}
\end{align}
The two global constraints in $R_{eg}$,
ranging from~\eqref{module:completion:constraint:e1} to~\eqref{module:completion:constraint:e4},
ensure that
if an examination task associated with a module $m$ is part of an examination plan,
then a valid combination of primary and secondary examination achieving module $m$
must also be a part of the examination plan,
with no further unnecessary examination tasks associated with the module being taken.
The first temporal constraint in $R_{et}$,
given in \eqref{examination:uniqueness},
forbids the same examination task from being completed in two distinct semesters.
Finally, the second temporal constraint
in~\eqref{dependency:constraint:e1} to~\eqref{dependency:constraint:e3},
implements the meaning of dependencies as presented when introducing the concept.

Last but not least,
an examination plan $(E_i)_{i=1}^n\subseteq (Ep \cup Es)^n$
for a basic study regulations problem with examination tasks $(\mathcal{B},\mathcal{E})$
is \emph{admissible} if
$\sebfunc((E_i)_{i=1}^{n})$ is a study plan for $\mathcal{B}$ and
$(E_i)_{i=1}^n \in\bigcap_{r\in R_{eg} \cup R_{et}} r$.

For illustration,
let us return to examination plan $(E_i)_{i=1}^4$.
We have already seen in Section~\ref{sec:approach} that $\sebfunc((E_i)_{i=1}^4)$
constitutes a valid study plan for the \emph{Cognitive Systems} program.
Similarly, the actual examination plan $(E_i)_{i=1}^4$ satisfies all constraints
in~\eqref{module:completion:constraint:e1} to~\eqref{dependency:constraint:e3},
which warrants its admissibility.
Unlike this, the aforementioned modified examination plan $E'$ fails to be admissible.
This is because,
first,  $R_t$, or more specifically~\eqref{cogsys:disjoint}, forces modules to occur at most once in a study plan and,
second, $R_{eg}$
forbids to have both $ep_{\mathit{bm}_1,1}$ and $ep_{\mathit{bm}_1,2}$ in $\overline{E'}$.
Hence, both $\sebfunc(E')$ and $E'$ violate constraints on study and examination plans, respectively.

\begin{figure}
  \begin{center}{\begin{picture}(300,120)(-150,-60)
        \put(-80,+40){\makebox(0,0){\framebox(80,20){$(E_j)_{j=1}^k$}}}
        \put(-80,-40){\makebox(0,0){\framebox(80,20){$(S_j)_{j=1}^k$}}}
        \put(+80,+40){\makebox(0,0){\framebox(80,20){$(E_i)_{i=1}^n$}}}
        \put(+80,-40){\makebox(0,0){\framebox(80,20){$(S_i)_{i=1}^n$}}}
        \put(-80,+30){\vector(0,-1){60}}
        \put(-40,-40){\vector(+1,0){80}}
        \put(+80,+30){\vector(0,-1){60}}
        \put(-40,+40){\vector(+1,0){80}}
        \put(-90,  0){\makebox(0,0){\sebfunc}}
        \put(+90,  0){\makebox(0,0){\sebfunc}}
        \put(   0,-55){\makebox(0,0){{Completing}}}
        \put(   0,+55){\makebox(0,0){{Completing}}}
      \end{picture}}
  \end{center}
  \caption{Relations between partial and total study and examination plans}
  \label{fig:map}
\end{figure}
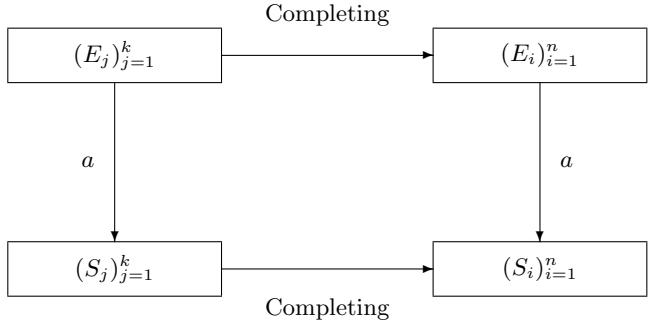

Although we focused so far on total study and examination plans,
in practice, a common use-case consists in completing partial ones.
The above formalization directly carries over to partial plans.
Even though we do not formally elaborate this,
we can show the relationships among partial and total study and examination plans
given in Figure~\ref{fig:map}.

\section{Encoding study regulations}
\label{sec:encodings}

In this section, we present an ASP-based approach
to represent study regulations and generate valid study plans.
We explain in detail the representation of basic study regulations,
and we discuss briefly its extension to examination tasks.\footnote{The complete encoding for study regulations with examination tasks is
available at \url{https://github.com/potassco/study-regulations/tree/v1.0.0}}
As usual, the representation is divided in two parts: a specific instance and a general encoding.
The instance represents the elements of a specific study regulation by a set of facts,
while the encoding provides the semantics associated with study regulations.
Given an instance that represents one study regulation, the answer sets of the encoding together with the instance
correspond to the study plans for that study regulation.

We try to keep the notation as close as possible
to the formalizations of the previous sections. We use the same symbols as before for sets and functions,
but always in lowercase, to adapt to the conventions of ASP.
For example, the sets $\overline{S}$, $S_i$ and $M_w$ are denoted by the terms
\lstinline{s}, \lstinline|s(i)| and \lstinline|m(w)|, respectively.
In what follows, we may use the logic programming notation to refer to
those sets and functions.

Listing~\ref{lst:cogsys:one} shows the first part of the instance \texttt{cogsys.lp}
for the \emph{Cognitive Systems} master,
that specifies the sets and functions of the study plan
(without examination tasks).
Sets are defined using atoms of the form \lstinline|in(e,a)|
that represent that the element \lstinline|e| belongs to the set \lstinline|a|.
For example, the atom \lstinline|in(bm1,b)| expresses that \lstinline|bm1| $\in$ \lstinline|b|.
Functions are defined similarly,
using atoms of the form \lstinline|map(f,e,v)|
that represent that the value of the function \lstinline|f|
applied to the element \lstinline|e| is \lstinline|v|.
For example, \lstinline|map(c,bm1,9)| expresses that
\mbox{\lstinline|c|$($\lstinline|bm1|$) = \ $\lstinline|9|}.
To define the facts more compactly,
we make extensive use of pooling using the operator `\lstinline{;}'.
For example, the three facts
\lstinline|in(bm1,b).|,
\lstinline|in(bm2,b).|, and
\lstinline|in(bm3,b).| defining the set \lstinline|b|
are captured by the single rule \lstinline|in((bm1;bm2;bm3),b).| in Line~\ref{lst:cogsys:b}.
\begin{lstlisting}[language=clingoht,label={lst:cogsys:one},linerange={1-20},basicstyle=\ttfamily\footnotesize,
  caption={First part of the instance of the Cognitive Systems master in \texttt{cogsys.lp}.}]
% b, f, a, o and p               % c                         |\label{lst:cogsys:start}|
in((bm1;bm2;bm3),b).             map(c,(bm1;bm2;bm3),9).|\label{lst:cogsys:b}|
in((fm1;fm2;fm3),f).             map(c,(fm1;fm2;fm3),6).
in((am11;am12;am21),a).          map(c,(am11;am12;am21),6).
in((am22;am31;am32),a).          map(c,(am22;am31;am32),6).
in(E,o) :- in(E,(f;a)).          map(c,(pm1;pm2;pm3),12). |\label{lst:cogsys:o}|
in((pm1;pm2;pm3),p).             map(c,im,15).
                                 map(c,msc,30).

% m                              % s
in(E,m) :- in(E,(b;f;a;p)).      map(s,bm1,w;s,bm2,s;s,bm3,w). |\label{lst:cogsys:m}|
in((im;msc),m).                  map(s,(fm1;fm2;fm3),w).
                                 map(s,(am11;am12;am21),e).
                                 map(s,(am22;am31;am32),e).
                                 map(s,(pm1;pm2;pm3),e).
                                 map(s,(im1;msc1),e).

% e                              % l and u
in(fm1,e).                       map(l,b,27;l,o,24;l,p,24;l,m,120).
                                 map(u,b,27;u,o,24;u,p,24;u,m,120).|\label{lst:cogsys:map:end}|
\end{lstlisting}

The definitions of the constraints in (\ref{cogsys:exogenous})-(\ref{cogsys:msc:pre})
provide the conditions that every study plan
$(S_i)_{i=1}^n\subseteq M^n$
must satisfy.
These  conditions
usually refer to operations over sets,
that we represent in ASP using prefix notation.
For example, the condition of (\ref{cogsys:exogenous}) refers to
the intersection of the sets $\overline{S}$ and $F$,
that is denoted in the logic program by the term
\lstinline|int(s,f)|.
Other terms can be used to denote the union, substraction and complement of sets.
The regulation constraints could in principle be very diverse,
but in our investigation of various study regulations we have
found that they can be captured by a few types of general constraints,
that we represent in ASP by different predicates.
The general encoding gives their semantics,
while the specific instance of each study regulation
provides facts over those predicates to represent the corresponding constraints.
Listing~\ref{lst:cogsys:two}
shows the second part of our example instance,
that specifies the constraints of the \emph{Cognitive Systems} master.
It uses atoms of the following form,
with the associated meaning
(where \lstinline|A| and \lstinline|B| denote sets,
 \lstinline|F| denotes a function, and
 \lstinline|L| and \lstinline|U| are integers):
\begin{itemize}
\item
\lstinline|empty(A)| means that \lstinline|A|$\ = \emptyset$, \item
\lstinline|equal(A,B)| means that \lstinline|A|$\ = \ $\lstinline|B|,
\item
\lstinline|subseteq(A,B)| means that \lstinline|A|$\ \subseteq \ $\lstinline|B|,
\item
\lstinline|card(A,leq,U)| means that \lstinline:|A|:$\ \leq \ $\lstinline|U|,
\item
\lstinline|sum(A,F,bw,(L,U))| means that \lstinline|L|$\leq \sum_{e \in \texttt{A}}$\lstinline|F|$(e) \leq$ \lstinline|U|, and
\item
\lstinline|sum(A,F,geq,L)| means that \lstinline|L|$\leq \sum_{e \in \texttt{A}}$\lstinline|F|$(e)$.
\end{itemize}
The general encoding includes more
predicates to represent other relations among sets,
like proper subset or superset,
and it also allows other types of comparisons within atoms of the
predicates \lstinline|card/3| and \lstinline|sum/4|.
Using these predicates,
the global constraints
(\ref{cogsys:exogenous})-(\ref{cogsys:msc})
are captured in Lines~\ref{lst:cogsys:gc1}-\ref{lst:cogsys:gc3}.
The first constraint consists of two conditions,
and this is accordingly represented by two facts.
Line~\ref{lst:cogsys:gc2}
uses pooling to refer to all the sets in
$\{$\lstinline|b|, \lstinline|o|, \lstinline|p|, \lstinline|m|$\}$,
and atoms over \lstinline|map/3| to capture the values
\lstinline|L| and \lstinline|U| of the functions \lstinline|l| and \lstinline|u|
applied to those sets.
The last rule of the block
defines a new set \lstinline|gc3| that consists of \lstinline|im| and \lstinline|msc|,
and compares it via \lstinline|subseteq| with \lstinline|s|.
Temporal constraints are represented in Lines~\ref{lst:cogsys:tc1} to~\ref{lst:cogsys:tc4}.
The first three use atoms of the predicate \lstinline|empty/1| that refer to
\lstinline|m(w)|, \lstinline|m(s)|,  and
the specific sets of modules \lstinline|s(i)| of each semester \lstinline|i|.
The last one defines the set \lstinline|tc4| that consists of \lstinline|msc|,
applies to it a new kind of temporal operator called \lstinline|before|,
and uses the resulting term in an atom of predicate \lstinline|sum/4|.
The term \lstinline|before(tc4)| denotes
the set of modules that occur in the study plan
before some element of \lstinline|tc4|;
in this case, before the module \lstinline|msc|.
Using our previous mathematical notation, the set \lstinline|before(tc4)| is
$\{ m \in M \ | \ m \in S_i \text{ and there is some  }m' \in \texttt{tc4}\cap S_k \text{ such that } i < k \}$.
The general encoding includes other similar operators like \lstinline|after| or \lstinline|between|.
\begin{lstlisting}[language=clingoht,label={lst:cogsys:two},linerange={22-36}, firstnumber=22,
  caption={Second part of the instance of the Cognitive Systems master in \texttt{cogsys.lp}.},basicstyle=\ttfamily\footnotesize]
% global constraints                                           |\label{lst:cogsys:cons:start}|
card(e,leq,2).  equal(int(s,f),e).|\label{lst:cogsys:gc1}|
sum(int(H,s),c,bw,(L,U)) :- H = (b;o;p;m), map(l,H,L), map(u,H,U).|\label{lst:cogsys:gc2}|
in(im,gc3). in(msc,gc3). subseteq(gc3,s).|\label{lst:cogsys:gc3}|

% temporal constraints
empty(int(s(I),s(J)))       :- I = 1..n, J = 1..n, I < J.|\label{lst:cogsys:tc1}|
empty(int(m(w),s(2*K  )))   :- K = 1..n, 1 <= 2*K,   2*K   <= n.|\label{lst:cogsys:tc2}|
empty(int(m(s),s(2*K-1)))   :- K = 1..n, 1 <= 2*K-1, 2*K-1 <= n.|\label{lst:cogsys:tc3}|
in(msc,tc4). sum(before(tc4),c,geq,90).|\label{lst:cogsys:end}||\label{lst:cogsys:tc4}|

\end{lstlisting}

Listing~\ref{lst:meta} shows the general encoding in \texttt{encoding.lp}.
It takes as input the constant \lstinline|n| that gives the length of the study plan.
This constant is used by the choice rule in Line~\ref{lst:meta:generate}
to generate the possible study plans,
represented by the sets \lstinline|s(i)|
for \lstinline|i| between \lstinline|1| and \lstinline|n|.
Then, Line~\ref{lst:meta:s} defines the set \lstinline|s|
as the union of all \lstinline|s(i)|'s,
and Line~\ref{lst:meta:mx} defines the sets \lstinline|m(w)|
and \lstinline|m(s)|.
After this, Lines~\ref{lst:meta:sets:start}-\ref{lst:meta:sets:end}
handle the additional sets that may occur in the constraints.
The first block of rules identifies the sets that occur as arguments in the constraints.
Then, the rules in Lines~\ref{lst:meta:int:set} and \ref{lst:meta:before:set} recursively
look for the sets occurring inside the operators \lstinline|int| and \lstinline|before|.
The encoding contains other similar rules for the
other operators, but we do not show them here.
Once all the new sets have been identified,
additional rules provide their definition.
Line~\ref{lst:meta:int:in} defines the intersection of two sets,
and Line~\ref{lst:meta:before:in} defines
the modules occurring before some module of another set.
The complete encoding includes further rules for the other operators.
The next part of the encoding, in Lines~\ref{lst:meta:cons:start}-\ref{lst:meta:cons:end},
enforces the constraints.
The first ones about \lstinline|empty/1|, \lstinline|subseteq/2| and \lstinline|equal/2|,
use the predicate \lstinline|in/2| to eliminate the cases that are not consistent with the constraints,
while those about \lstinline|card/3| and \lstinline|sum/4| rely on
cardinality and aggregate atoms for that task.
For example, the condition
\lstinline:|A|:$\ \leq \ $\lstinline|U|
for
\lstinline|card(A,leq,U)| is captured by
the cardinality atom \lstinline|{ in(E,A) } U|,
and the condition
\mbox{\lstinline|L|$\leq \sum_{e \in \texttt{A}}$\lstinline|F|$(e) \leq$ \lstinline|U|} for
\lstinline|sum(A,F,bw,(L,U))| is captured by
the aggregate atom \lstinline|L #sum{ V,E : in(E,A), map(F,E,V) } U|.
Finally, the last block of statements in Lines~\ref{lst:meta:show:start} and~\ref{lst:meta:show:end}
displays the sets \lstinline|s(i)|.
\begin{lstlisting}[language=clingoht,label={lst:meta},caption={Encoding for all study regulations in \texttt{encoding.lp}.},basicstyle=\ttfamily\footnotesize]
% generate
{ in(E,s(I)) } :- in(E,m), I=1..n.|\label{lst:meta:generate}|

% s = s(1) U ... U s(n)
in(E,s) :- in(E,s(I)).|\label{lst:meta:s}|

% m(w) and m(s)
in(E,m(X)) :- X = (s;w), in(E,m), map(s,E,X).|\label{lst:meta:mx}|

% additional sets
set(A) :-      empty(A).|\label{lst:meta:sets:start}|
set(A) :- subseteq(A,B).    set(A) :-   equal(A,B).
set(B) :- subseteq(A,B).    set(B) :-   equal(A,B).
set(A) :-   card(A,R,L).    set(A) :- sum(A,M,R,L).
%
set(A) :- set( int(A,B)).   set(B) :- set(int(A,B)).|\label{lst:meta:int:set}|
set(A) :- set(before(A)).|\label{lst:meta:before:set}|
%
in( E, int(A,B)) :- set( int(A,B)), in(E,A), in(E,B).|\label{lst:meta:int:in}|
in(E1,before(A)) :- set(before(A)), in(E1,s(I)), in(E2,A), in(E2,s(J)), I < J.|\label{lst:meta:before:in}||\label{lst:meta:sets:end}|

% constraints
:- empty(A), in(E,A).|\label{lst:meta:cons:start}|
%
:- subseteq(A,B), in(E,A), not in(E,B).
%
:- equal(A,B), in(E,A), not in(E,B).
:- equal(A,B), not in(E,A), in(E,B).
%
:- card(A,leq,U), not { in(E,A) } U.|\label{lst:meta:cons:card}|
%
:- sum(A,F,bw,(L,U)), not L #sum{ V,E : in(E,A), map(F,E,V) } U.
:- sum(A,F,geq,   L), not L #sum{ V,E : in(E,A), map(F,E,V) }.|\label{lst:meta:cons:end}|

% display
#show.                   |\label{lst:meta:show:start}|
#show (M,I) : in(M,s(I)).|\label{lst:meta:show:end}|
\end{lstlisting}

We can now run the ASP solver \clingo\
with the instance for the \emph{Cognitive Systems} master
and the general encoding.
For \lstinline|n=4| we obtain, among others, an answer set that corresponds to the admissible study plan
$S$ of Example~\ref{sec:example:cogsys}:
\begin{lstlisting}[basicstyle=\small\ttfamily,numbers=none,xleftmargin=\parindent]
clingo -c n=4 cogsys.lp encoding.lp
...
Answer: 1
(bm1,1) (bm3,1) (fm1,1) (am12,1) (bm2,2) (am21,2) (pm1,2)
(im,3) (am31,3) (pm3,3) (msc,4)
\end{lstlisting}
This problem is solved in less than a second,
but we have not evaluated the
scalability of our approach yet.
Note also that our current encodings are designed for readability,
but we foresee that they can be made more efficient.

The extension to handle examination tasks is not involved.
The first part of the instance is extended by
the specification of the sets $E_p$, $E_s$ and $D$
using predicate \lstinline|in/2|,
and of the functions $e_p$ and $e_s$
using predicate \lstinline|map/3|.
The second part specifies the constraints in $R_{eg}$ and $R_{eg}$.
As an example, the first global constraint is represented by the rule \begin{lstlisting}[basicstyle=\small\ttfamily,numbers=none,xleftmargin=\parindent]
implies(
  neg(empty(int(ee,expand(union(EP,ES))))),
  in'(int(ee,expand(EP)),EP)
) :- in(M,m), map(ep,M,EP), map(es,M,ES).
\end{lstlisting}
Looking at the body,
variable \lstinline|M| represents a module,
\lstinline|EP| denotes $\mathit{e_p}(\texttt{M})$, and
\lstinline|ES| denotes $\mathit{e_s}(\texttt{M})$.
In the second line,
the term \lstinline|ee| refers to the set $\overline{E}$,
defined in the extended general encoding,
while the set operator \lstinline|expand| takes one set of sets
and returns the union of the elements of that set.
In the third line, relation
\lstinline|in'| is a version of \lstinline|in|
where the first argument is a set and the second is a set of sets.
With this, the second line refers to the set
$\overline{E}\cap\overline{\mathit{e_p}(\texttt{M})\cup\mathit{e_s}(\texttt{M})}$
and checks whether that set is not empty,
and the third line checks if
$\overline{E}\cap\overline{\mathit{e_p}(\texttt{M})}\in\mathit{e_p}(\texttt{M})$.
All together, the rule states that for every module \lstinline|M|
the condition of the second line implies the condition of the third one,
just like the
constraint~(\ref{module:completion:constraint:e1})-(\ref{module:completion:constraint:e2}).

The general encoding for examination tasks
is extended accordingly to accomodate
the new set operations, like \lstinline|expand|,
and the logical connectives,
like \lstinline|implies| or \lstinline|neg|.
The main change of the encoding takes place at the generation part,
where the choice rule of Line~\ref{lst:meta:generate}
is replaced by
another choice rule that generates possible examination plans
together with a normal rule that implements function $a$ and defines the corresponding possible study plans.

 \section{ASP-driven user interface}
\label{sec:ui}

In this section, we sketch our interactive prototypical User Interface (UI)
for creating study plans in accordance with study regulations.
Notably,
the UI is generated and driven by ASP, more precisely the \clinguin\ system.\footnote{\url{https://clinguin.readthedocs.io/en/latest}}

\begin{figure}[ht]
    \centering
    \frame{\includegraphics[scale=0.12]{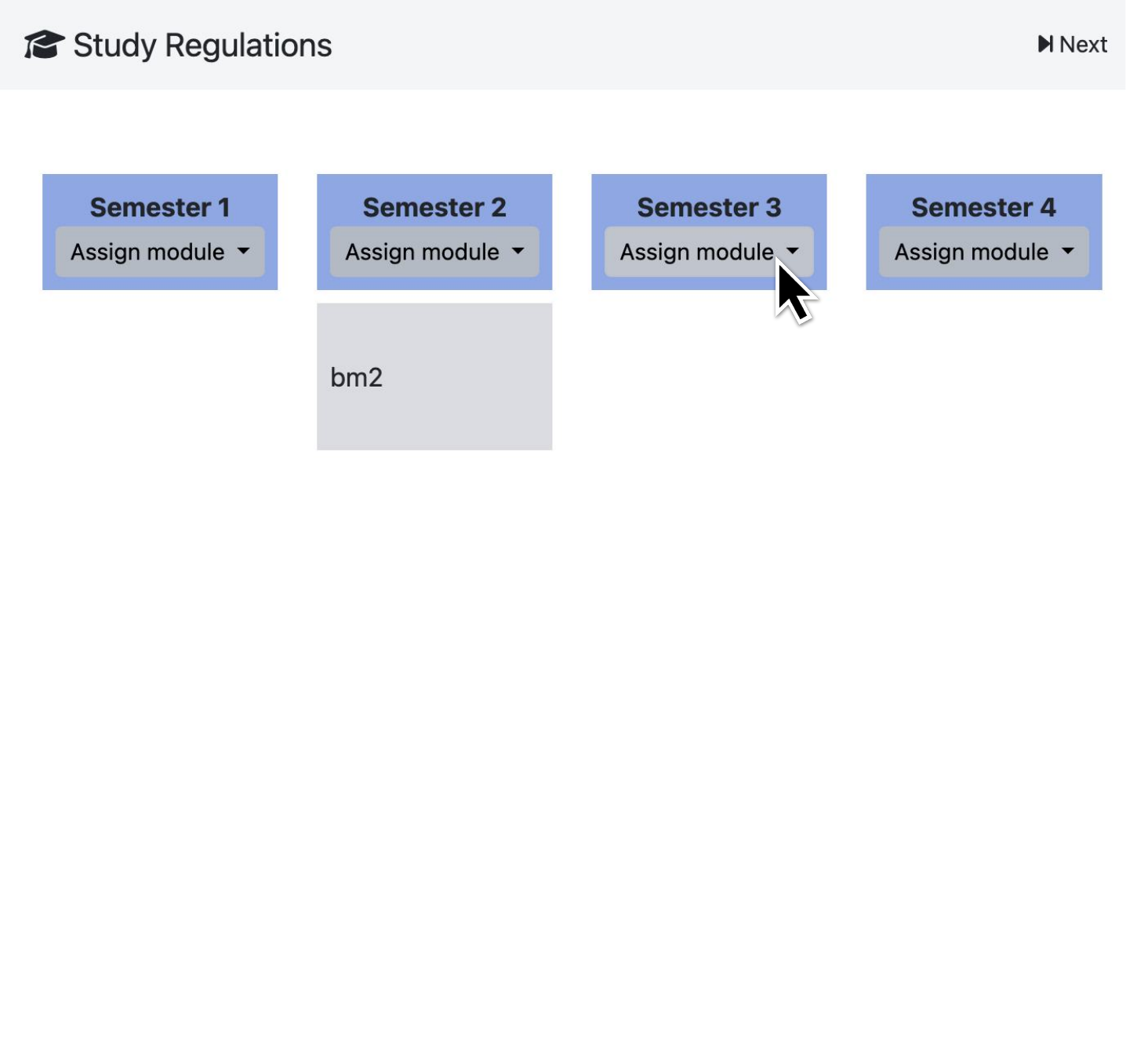}}
    \frame{\includegraphics[scale=0.12]{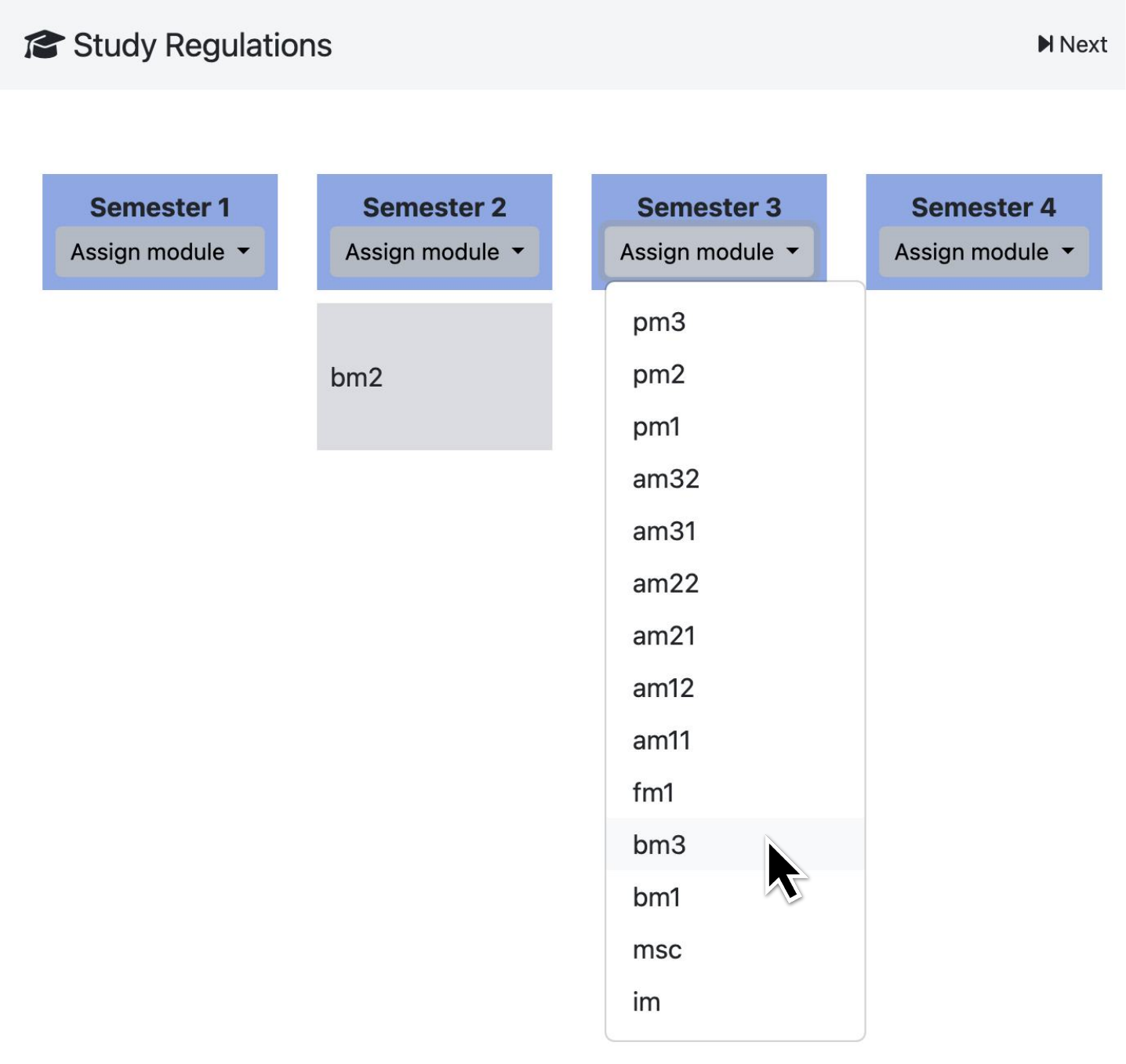}}
    \frame{\includegraphics[scale=0.12]{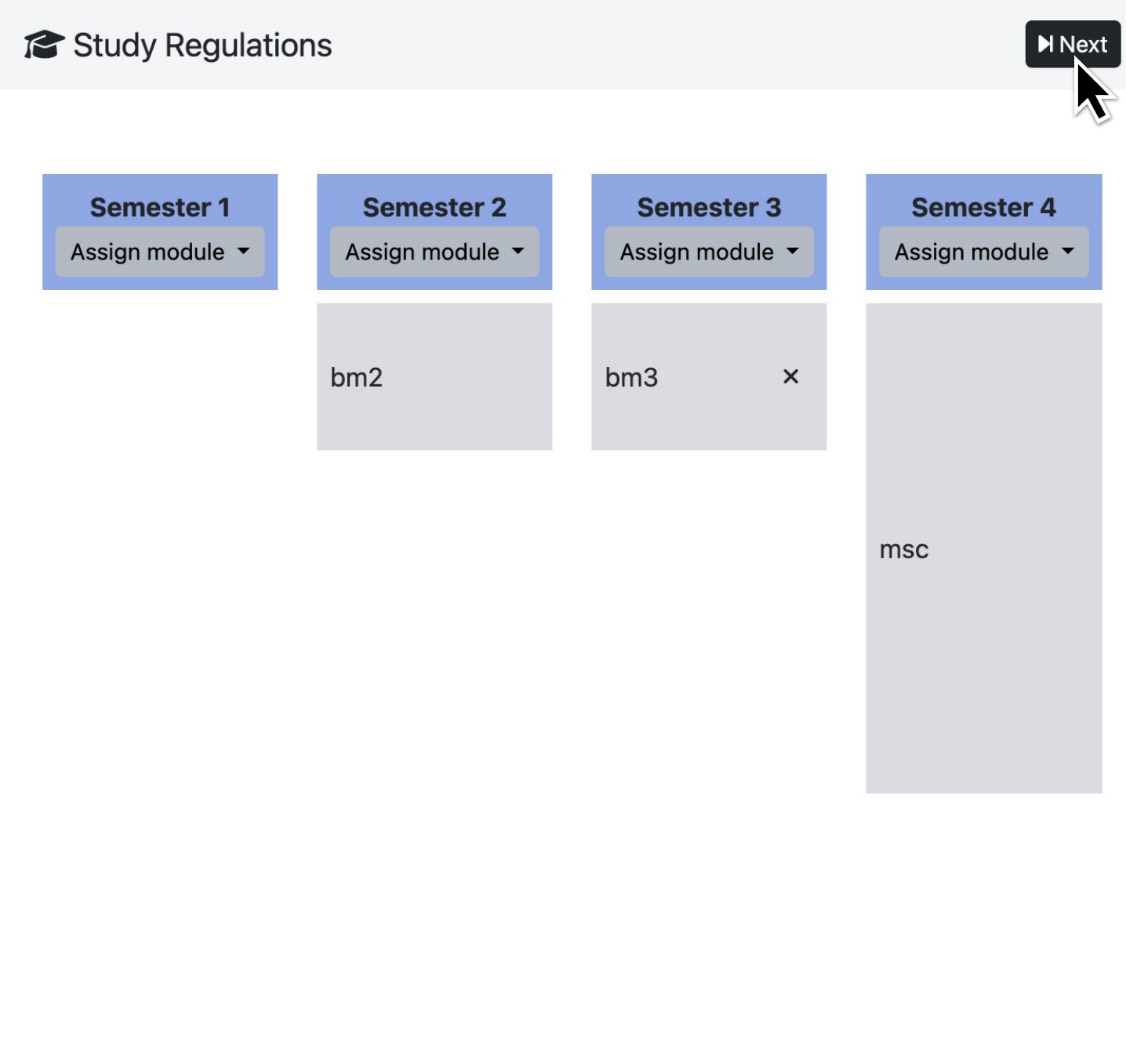}}
    \frame{\includegraphics[scale=0.12]{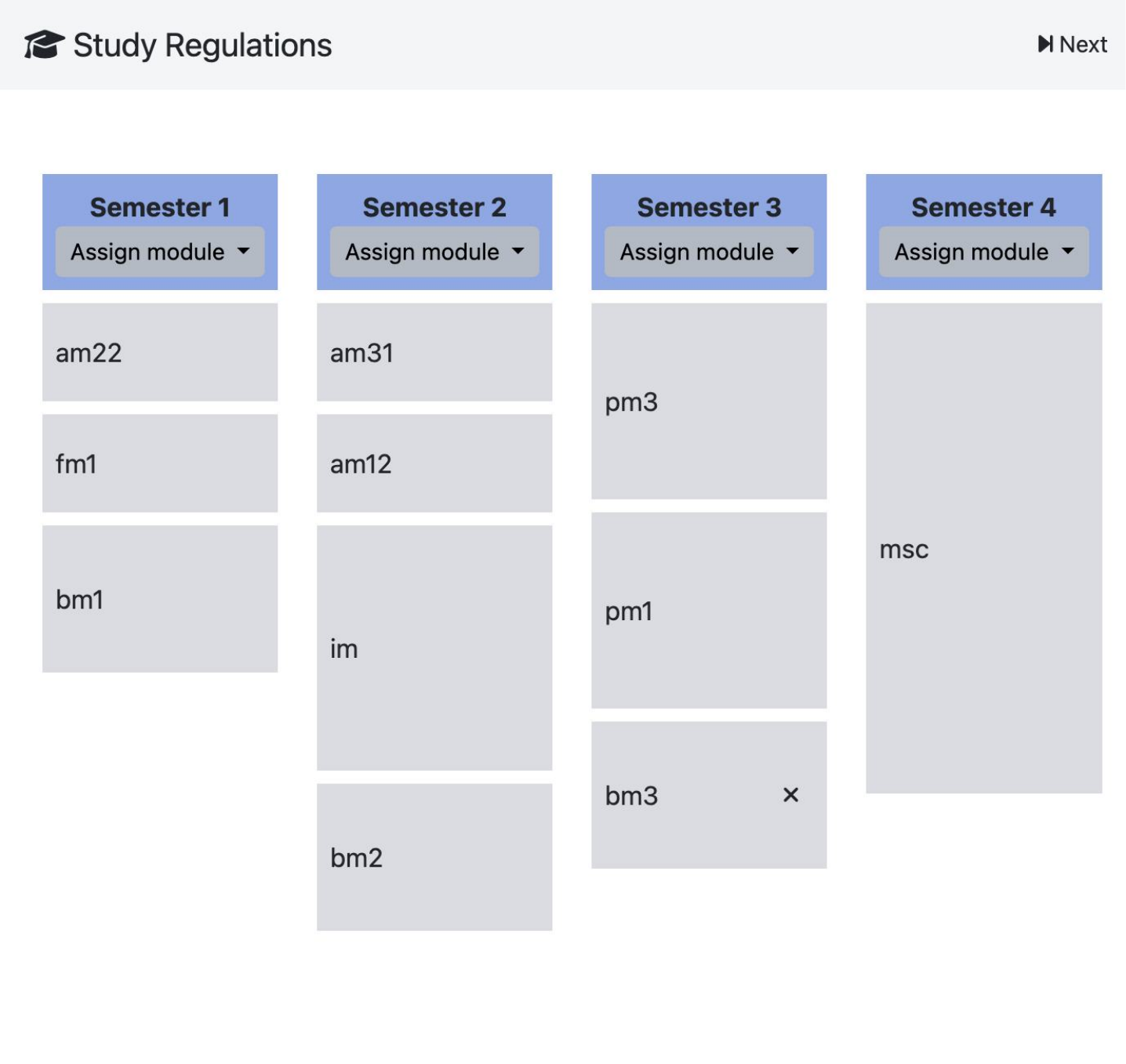}}
    \caption{User interaction via mouse actions in \clinguin.}
    \label{fig:ui}
\end{figure}
A detailed description of the UI can be found in our preliminary publication,
where \textit{tkinter} is used for rendering.
For a complement, we concentrate in what follows on the modern web-based front-end \emph{Angular}.\footnote{\url{https://angular.io}}$^,$\footnote{A detailed account of \clinguin\ has also been submitted to ICLP'24.}
To this end,
\clinguin\ uses a few dedicated predicates to define UIs and the treatment of user-triggered events.
This simple design greatly facilitates the specification of continuous user interactions with an ASP system,
in our case \clingo~\citep{karoscwa21a}.
More precisely,
the UI is defined by predicates \lstinline|elem/3|, \lstinline|attr/3| and \lstinline|when/4|,
for specifying the UI's layout, style and functionality, respectively.

\begin{lstlisting}[float=ht,caption={An encoding for the prototype UI (\texttt{ui.lp})},label={lst:ui},numbers=left,language=clingos]
elem(w, window, root).%%#(\label{ui:elem:w}#)

elem(s(I), container, w) :- semester(I).%%#(\label{ui:elem:s}#)

elem(s_t(I), container, s(I)) :- semester(I).%%#(\label{ui:elem:st}#)
attr(s_t(I), class, ("bg-primary";"bg-opacity-50"; %%#(\label{ui:class:st}#)
                     "fw-bold";"p-2";"m-1"       )) :- semester(I).

elem(s_dd(I), dropdown_menu, s_t(I)) :- semester(I).%%#(\label{ui:elem:sdd}#)
attr(s_dd(I), selected, "Assign module") :- semester(I).%%#(\label{ui:select:sdd}#)

possible_module(I,E) :- _any(in(E,s(I))), not _all(in(E,s(I))).%%#(\label{ui:pos}#)
elem(s_ddi(I,E), dropdown_menu_item, s_dd(I)) :- possible_module(I,E).%%#(\label{ui:elem:ddmi}#)
attr(s_ddi(I,E), label, E) :- possible_module(I,E).%%#(\label{ui:label:ddmi}#)
when(s_ddi(I,E), click, call, add_assumption(in(E,s(I)))) :- %%#(\label{ui:call:ddmi}#)
            possible_module(I,E).

assigned_module(I,E) :- _all(in(E,s(I))).%%#(\label{ui:show}#)
assigned_module(I,E) :- in(E,s(I)),_clinguin_browsing. %%#(\label{ui:showb}#)
elem(s_module(I,E), container, s_modules(I)) :- assigned_module(I,E).%%#(\label{ui:elem:sm}#)
attr(s_module(I,E), height, C*10) :- assigned_module(I,E), map(c,E,C).%%#(\label{ui:height:sm}#)
\end{lstlisting}
We show in Listing~\ref{lst:ui} the relevant sections of the encoding used to generate the UI snapshots in Figure~\ref{fig:ui}.\footnote{The full encoding can be found in \url{https://github.com/potassco/study-regulations/tree/v1.0.0}}
This encoding is passed along with our study regulations encoding from Section~\ref{sec:encodings} to \clinguin,
which allows for browsing and completing (partial) study plans.

Line~\ref{ui:elem:w} creates a window element.
Line~\ref{ui:elem:s} creates a container for each semester, which is placed in the window.
Lines~\ref{ui:elem:st} and~\ref{ui:class:st} define the title container for each semester,
assigning values to the attribute \lstinline{class} to define the blue background, bold font, padding and margin.
Lines~\ref{ui:elem:sdd} and~\ref{ui:select:sdd} define a dropdown menu with the text `Assign module'.

The possible modules to be assigned are defined in predicate \lstinline{possible_module/1} on Line~\ref{ui:pos},
using the assignments that appear in any model (union) but are not in all models (intersection),
via predicates \lstinline|_any/1| and \lstinline|_all/1|, respectively.
This predicate is used in Lines~\ref{ui:elem:ddmi} to \ref{ui:call:ddmi} for defining the items in the dropdown menu.

In the second snapshot of Figure~\ref{fig:ui}, we notice that module $\mathit{bm}_2$ does not appear in the options
since there is no model where this module is assigned to the third semester.
When the module $\mathit{bm}_3$ is clicked, Line~\ref{ui:call:ddmi} adds the corresponding atom
as an assumption~\citep{ankamasc12a,aldofiprri23a}.
Lines~\ref{ui:show} and~\ref{ui:showb} define the modules that are assigned to a semester (gray boxes).
These can be the ones appearing in all models (Line~\ref{ui:show}) due to a user selection or inference,
or the ones in the current model when the user is browsing solutions,
as indicated by atom \lstinline{_clinguin_browsing}.

Predicate \lstinline{assigned_modules/2} is then used in Lines~\ref{ui:elem:sm} and~\ref{ui:height:sm} to create
the corresponding container, using the number of credits of the module to define the height in Line~\ref{ui:height:sm}.
The third snapshot shows module `bm3' assigned to the third semester after the previous click.
Modules selected by the user include a button marked with `x' to remove the selection;
unlike assignments forced by the encoding, such as module `msc' in the fourth semester.
Upon clicking in the `Next' button, the last snapshot shows one full study plan consistent with the user's selections.

 \section{Related work}\label{sec:related-work}

ASP, event calculus and process mining techniques were already used in \citep{miherojudogolasc23a} for solving study regulation problems.
However, \cite{miherojudogolasc23a} presents an overview of the \emph{AIStudyBuddy} project, and contains neither a
formalization of study regulations nor any implementation details.
Unlike this, \cite{sagume23a} presents a web-based Decision Support System for a degree planning problem along with
a mathematical formalization.
Degree requirements are mentioned but not formalized.
Also, \cite{bayani23a} considers educational planning problems.
This includes stress and learning effects on students in a personalized study plan generation.
It aims at reducing the duration of student plans and is implemented via integer linear programming.
\cite{sclewi18a} present a curriculum timetable validation tool by modeling constraints in language~\emph{B}.
Finally,
\cite{shlili22a} present a data-driven approach for implementing a course recommendation algorithm.
A traditional collaborative filtering algorithm is extended to consider additional course path data.

 \section{Summary}\label{sec:discussion}

We have introduced a conceptualization of study regulations based on set-based constraints.
This formalization is both simple and general to capture a wide range of study regulations.
We indicated how the basic formulation easily extends to more complex constructions.
This will be further elaborated in future work.
The identification of basic principles in study regulations also allowed us to obtain a very general ASP encoding.
The building blocks of each study regulation are captured in terms of facts so that the actual encoding is
also applicable to a large range of study programs.
Finally, we have described an ASP-driven user interface for interactive elaboration of study plans.
Again, the interface is designed in a generic way and broadly applicable.
Moreover, this case study serves as a nice illustration of \clinguin\ and how it can be used for interactive
ASP applications.

Finally, study regulations offer a very rich playground for applications of knowledge representation and
reasoning techniques.
Study plans have a light temporal flavor and resemble finite traces in linear temporal logic~\citep{giavar13a}.
The creation of a study plans amounts to a configuration task, which also brings about interaction and
explainability.
Finally, we have so far only been concerned with the hard constraints emerging from study regulations but there
is so much commonsense knowledge involved, like defaults, preferences, deontic laws, updates, etc.

\paragraph{Acknowledgments.}
This work was partly funded by DFG grant SCHA 550/15 and BMBF project \textit{CAVAS+} (16DHBKI024).

\bibliographystyle{tlplike}

\end{document}